\documentclass[sigconf]{acmart}

\AtBeginDocument{%
  \providecommand\BibTeX{{%
    \normalfont B\kern-0.5em{\scshape i\kern-0.25em b}\kern-0.8em\TeX}}}
    
\usepackage{multicol,multirow}
\usepackage[normalem]{ulem}

\copyrightyear{2023}
\acmYear{2023}
\setcopyright{rightsretained}
\acmConference[FashionXRecSys '22]{Sixteenth ACM Conference on Recommender Systems}{September 18--23, 2022}{Seattle, WA, USA}
\acmBooktitle{Sixteenth ACM Conference on Recommender Systems (FashionXRecSys '22), September 18--23, 2022, Seattle, WA, USA}
\acmDOI{10.1007/978-3-031-22192-7_6}
\acmISBN{978-3-031-22192-7}

\begin{document}

\title{A Dataset for Learning Graph Representations to Predict Customer Returns in Fashion Retail}

\author{Jamie McGowan}
\email{j.mcgowan.18@ucl.ac.uk}
\orcid{0000-0003-3502-8719}
\affiliation{%
  \institution{University College London}
  \city{London}
  \country{UK}
}

\author{Elizabeth Guest}
\email{elizabeth.guest.21@ucl.ac.uk}
\affiliation{%
  \institution{University College London}
  \city{London}
  \country{UK}
}

\author{Ziyang Yan}
\email{ziyang.yan.17@ucl.ac.uk}
\affiliation{%
  \institution{University College London}
  \city{London}
  \country{UK}
}

\author{Zheng Cong}
\email{zheng.cong.20@ucl.ac.uk}
\affiliation{%
  \institution{University College London}
  \city{London}
  \country{UK}
}

\author{Neha Patel}
\email{neha.patel@asos.com}
\affiliation{%
  \institution{ASOS AI}
  \city{London}
  \country{UK}
}

\author{Mason Cusack}
\email{mason.cusack@asos.com}
\affiliation{%
  \institution{ASOS AI}
  \city{London}
  \country{UK}
}

\author{Charlie Donaldson}
\email{charlie.donaldson@asos.com}
\affiliation{%
  \institution{ASOS AI}
  \city{London}
  \country{UK}
}

\author{Sofie de Cnudde}
\email{sofiede.cnudde@asos.com}
\affiliation{%
  \institution{ASOS AI}
  \city{London}
  \country{UK}
}

\author{Gabriel Facini}
\email{g.facini@ucl.ac.uk}
\affiliation{%
  \institution{University College London}
  \city{London}
  \country{UK}
}

\author{Fabon Dzogang}
\email{fabon.dzogang@asos.com}
\affiliation{%
  \institution{ASOS AI}
  \city{London}
  \country{UK}
}

\renewcommand{\shortauthors}{McGowan, et al.}

\begin{abstract}
  We present a novel dataset collected by ASOS (a major online fashion retailer) to address the challenge of predicting customer returns in a fashion retail ecosystem. With the release of this substantial dataset we hope to motivate further collaboration between research communities and the fashion industry. We first explore the structure of this dataset with a focus on the application of Graph Representation Learning in order to exploit the natural data structure and provide statistical insights into particular features within the data. In addition to this, we show examples of a return prediction classification task with a selection of baseline models (i.e. with no intermediate representation learning step) and a graph representation based model.
  We show that in a downstream return prediction classification task, an F1-score of 0.792 can be found using a Graph Neural Network (GNN), improving upon other models discussed in this work. Alongside this increased F1-score, we also present a lower cross-entropy loss by recasting the data into a graph structure, indicating more robust predictions from a GNN based solution. These results provide evidence that GNNs could provide more impactful and usable classifications than other baseline models on the presented dataset and with this motivation, we hope to encourage further research into graph-based approaches using the ASOS GraphReturns dataset.
\end{abstract}




\maketitle

\section{Introduction}\label{sec:intro}
Part of the unique digital experience that many fashion retailers deliver is the option to return products at a small or no cost to the customer. However, unnecessary shipping of products back and forth incurs a financial and environmental cost. With many fashion retailers having a commitment to minimizing the impact of the fashion industry on the planet, providing a service which can forecast returns and advise a customer of this at purchase time is in line with these goals.

With the continual development of e-commerce platforms, it is important that systems are able to model the user's preferences within the platform's ecosystem by using the available data to guide users and shape the modern customer experience. One approach to this challenge, which has sparked huge interest in the field of recommendation systems~\cite{Wu:GNN}, are representation learning based methods. Representation learning provides a framework for learning and encoding complex patterns present in data, which more naive machine learning (ML) approaches are unable to capture as easily. However at present, the available data that is able to facilitate such research avenues is scarce. Further to this, the number of available datasets which include anonymised customer and product information (and their interactions) is even less available.

E-commerce platforms in the fashion industry are in a unique position to contribute to this research by making data publicly available for use by the machine learning community. Of particular interest to ASOS is the application of machine learning to predicting customer returns at purchase time, due to this, we present the ASOS GraphReturns dataset in this article. The labelled purchase (return or not returned) connections between customers and products in this dataset naturally lends itself to a graph structure which has motivated our interest in encouraging the exploration of graph representation learning based solutions, which we provide an example of in Sect.~\ref{sec:baseline-results}. Graph Neural Networks (GNNs) have been the subject of immense success in recent years~\cite{jumper2021highly,stokes2020:antibiotic,sanchez2020:physics,lange2020:traffic,eksombatchai2018:pixie} and provide an intuitive way to exploit structured data. Another benefit of using GNNs is that they are able to make predictions for new instances not seen before. This is a particular attractive feature for industry environments where new products and customers are continually added.

In this work, we first present the ASOS GraphReturns dataset\footnote{The dataset can be found at \href{https://osf.io/c793h/}{https://osf.io/c793h/}.} and discuss some of the properties and features of this data. Using this data we then provide some examples demonstrating the use of GNNs with this data based on the downstream task of predicting customer returns. This information may then be used to inform customers based on their choice and make a personalised recommendation (i.e. a different size, style, colour etc.) at purchase time that has a lower probability of being returned.

The structure of the document is as follows: Sect.~\ref{sec:data} describes the novel fashion retail dataset, Sect.~\ref{sec:method} overviews the methodology and some example benchmark results are discussed in Sect.~\ref{sec:baseline-results}. Finally in Sect.~\ref{sec:conclusion} we summarise this contribution and provide some insights into potential further studies which could benefit from this dataset.

\section{Data Description}\label{sec:data}
\begin{figure*}[t]
    \centering
    \includegraphics[width=\linewidth]{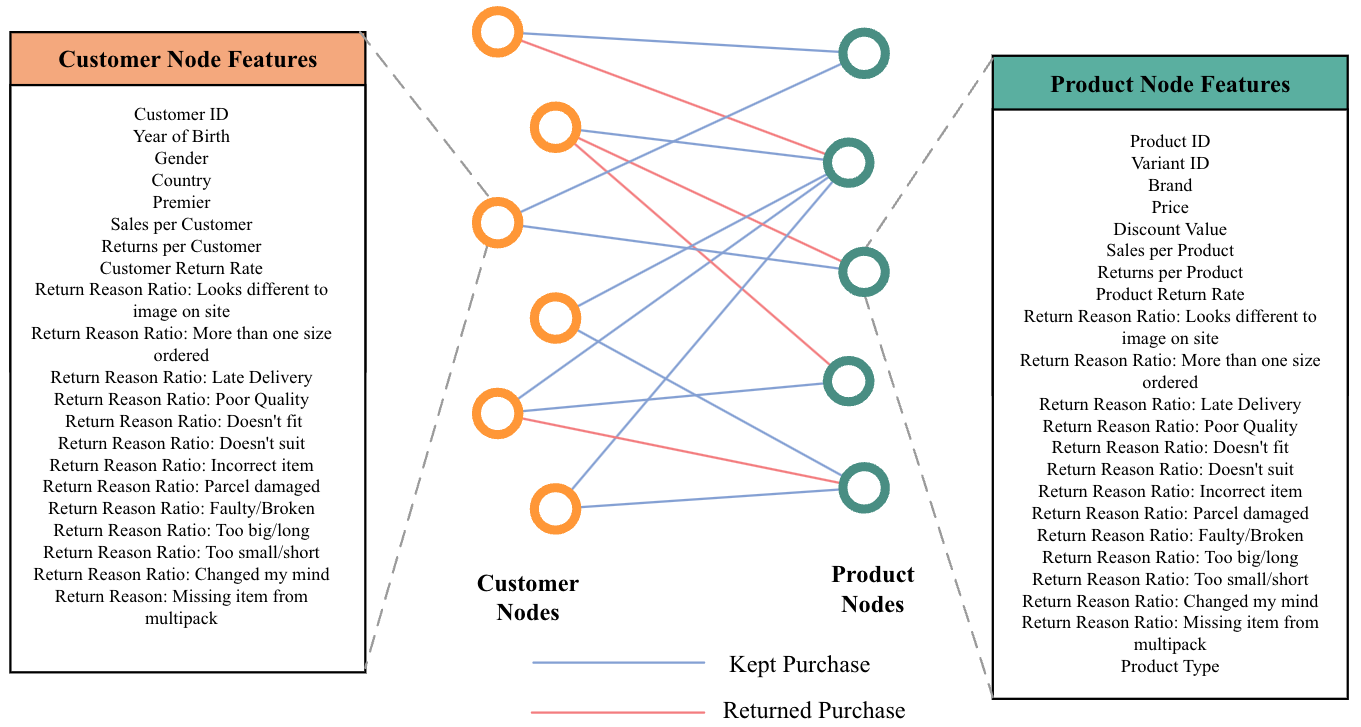}
    \caption{The raw data structure includes customer and product specific information linked by purchases. These purchase links are labeled with a no return (blue) or return (red) label. The entire list of node features for customers and products is also provided here.}
    \label{fig:raw_data_structure}
\end{figure*}

The train (test) data contains purchases and returns recorded by ASOS between Sept-Oct 2021 (Oct-Nov 2021), including the corresponding anonymous customer and product variant\footnote{Note that product variants include variations in size and colour and therefore a product may contain multiple variants.} specific information. The data is organised into customers (with hashed customer ID's to preserve anonymity), product variants and events (i.e. a purchase or return of a product by a customer). The training (testing) dataset includes $\sim770,000$ ($\sim820,000$) unique customers and $\sim410,000$ ($\sim410,000$) product variants, where every customer has at least one return and each product variant has been purchased at least once. To connect customers and products the data contains a total of 1.4M (1.5M) purchase events each labeled as a return (1) or no return (0) in both the training and testing datasets. The problem of predicting customer returns is then presented as an edge classification task as depicted in Fig.~\ref{fig:raw_data_structure}. This structure is similar to that of the Amazon reviews data~\cite{Amazon:data} which also includes labeled links between customers and products.

Within each customer/product variant node, we also include specific node features, such as the average return rate, the ratios of different reasons for returns, and historical information relating to the number of purchases/returns made. Fig.~\ref{fig:raw_data_structure} displays an exhaustive list of all the features included in this dataset.
\begin{figure*}[t]
\begin{minipage}[c]{0.45\textwidth}
    \centering
    \includegraphics[width=\linewidth]{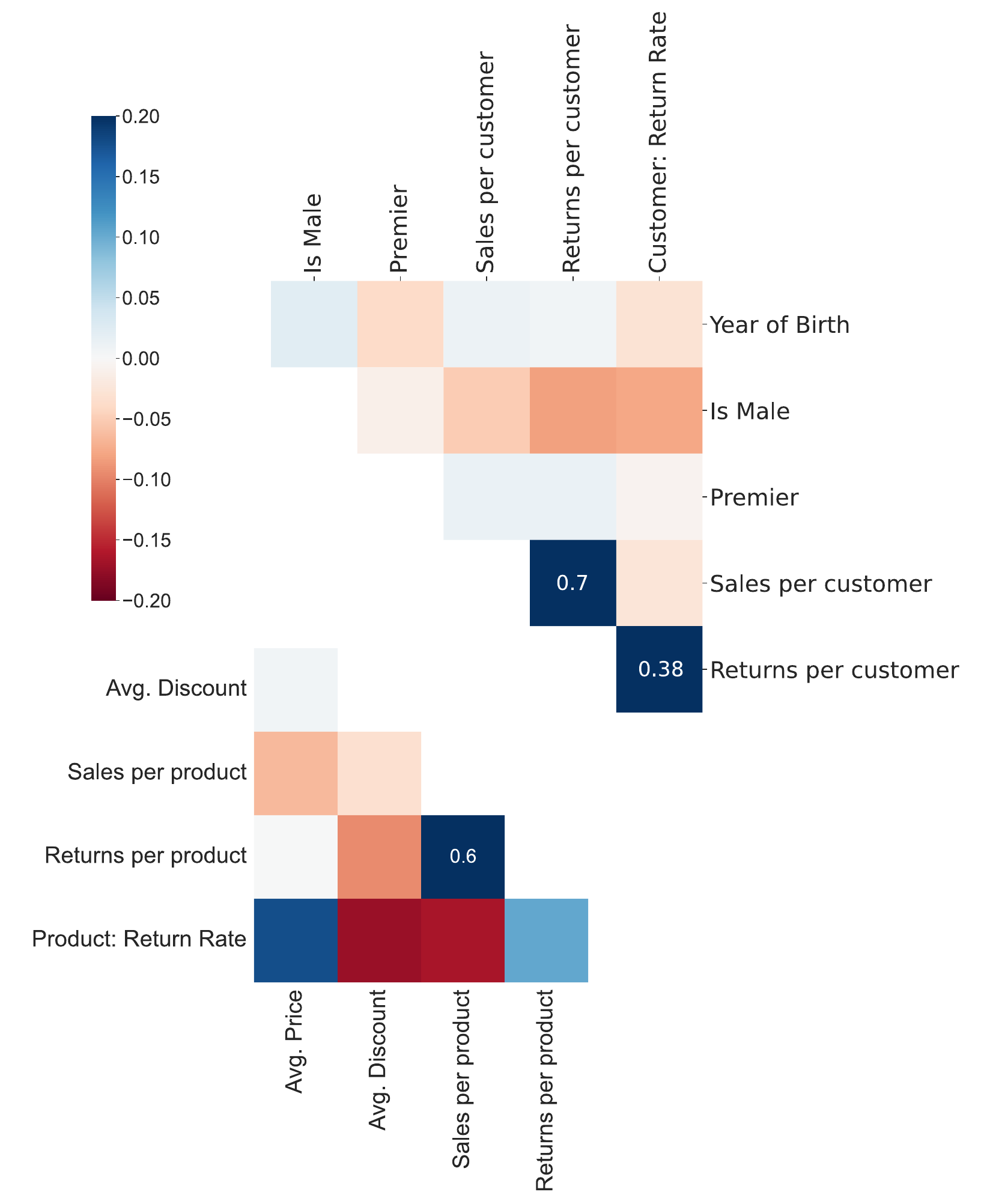}
\end{minipage}
\begin{minipage}[c]{0.5\textwidth}
    \includegraphics[width=\linewidth]{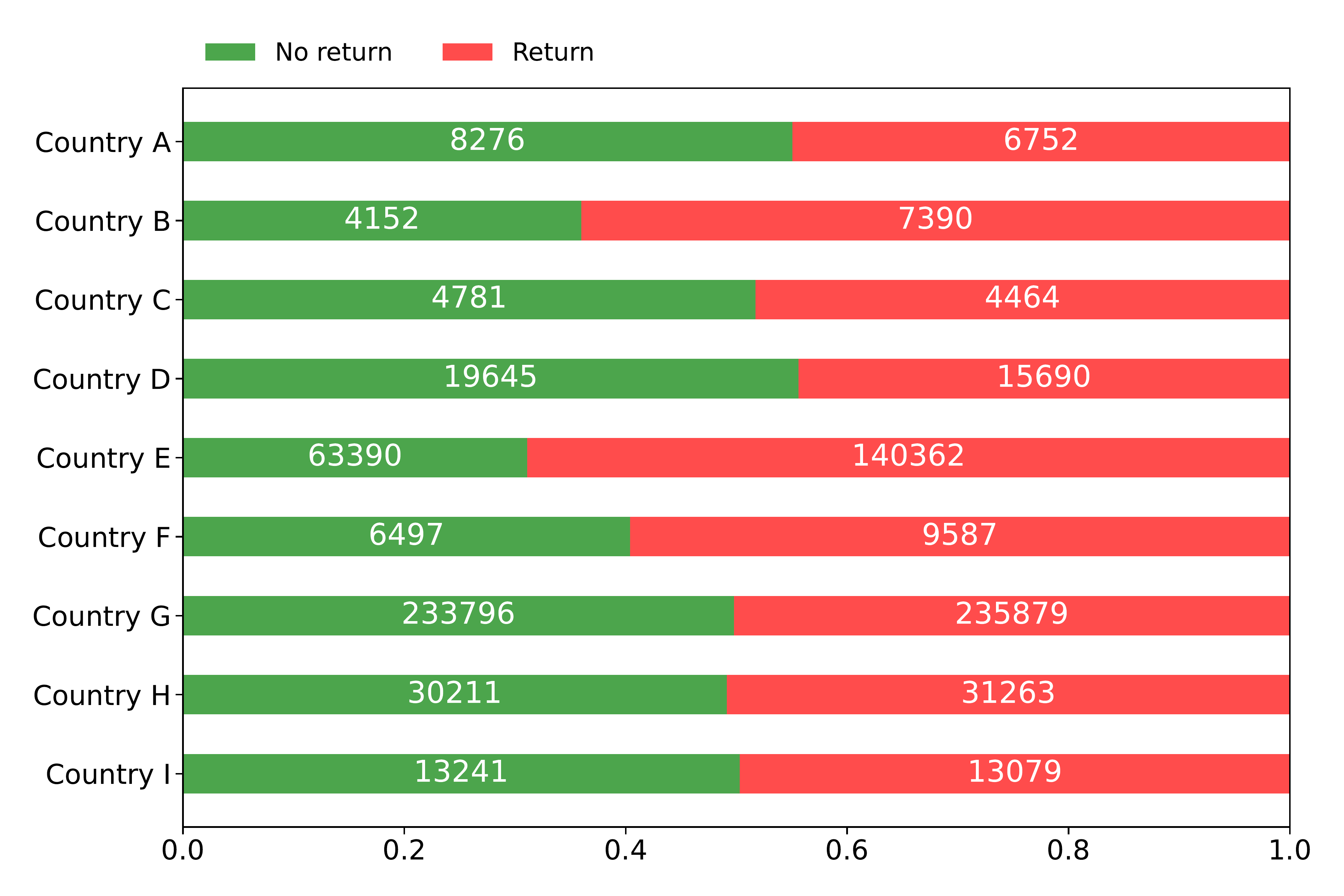}
    \includegraphics[width=\linewidth]{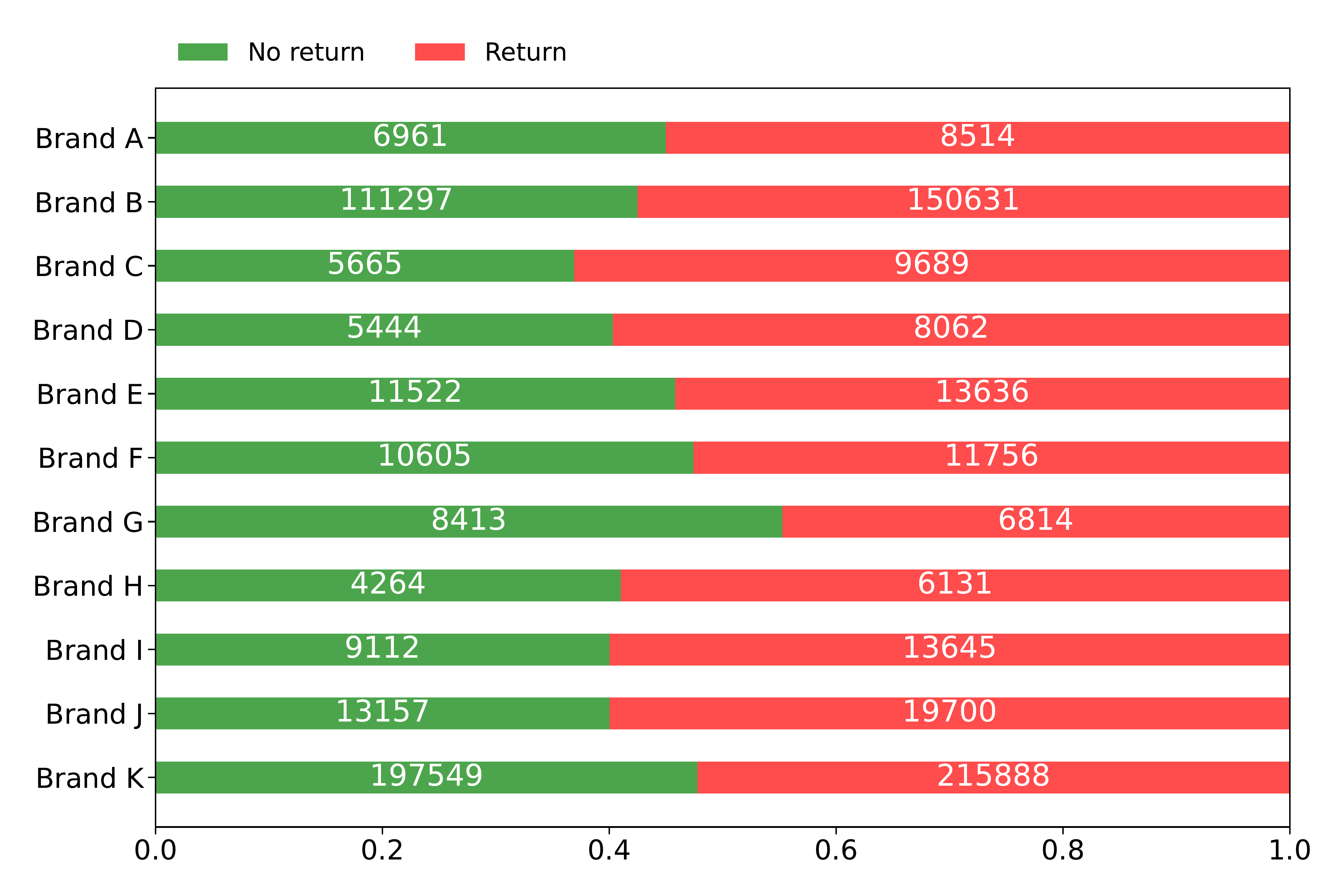}
\end{minipage}
    \caption{General summary of data statistics including correlations between customer and product specific features (left) and distributions of return labels (right) within each country (top) and brand (bottom).}
    \label{fig:data_corr_example}
\end{figure*}
Fig.~\ref{fig:data_corr_example} (left) displays a subset of correlations between customer (top) and product (bottom) features. Within these correlations, one can observe strong associations such as male customers being less likely to make a return or a more expensive product in general having a higher return rate. Fig.~\ref{fig:data_corr_example} (right) summarises a selection of statistics related to the distribution of return labels across countries and brands included within the data. It can be seen that the data shows a larger proportion of returns across specific individual markets which could prove useful in ML based classification tasks\footnote{Due to the manner in which this dataset is constructed (i.e. only including customers who have at least one return), these statistics do not reflect the true ASOS purchase/return statistics.}.

\begin{figure*}[t]
    \centering
    \includegraphics[width=0.8\linewidth]{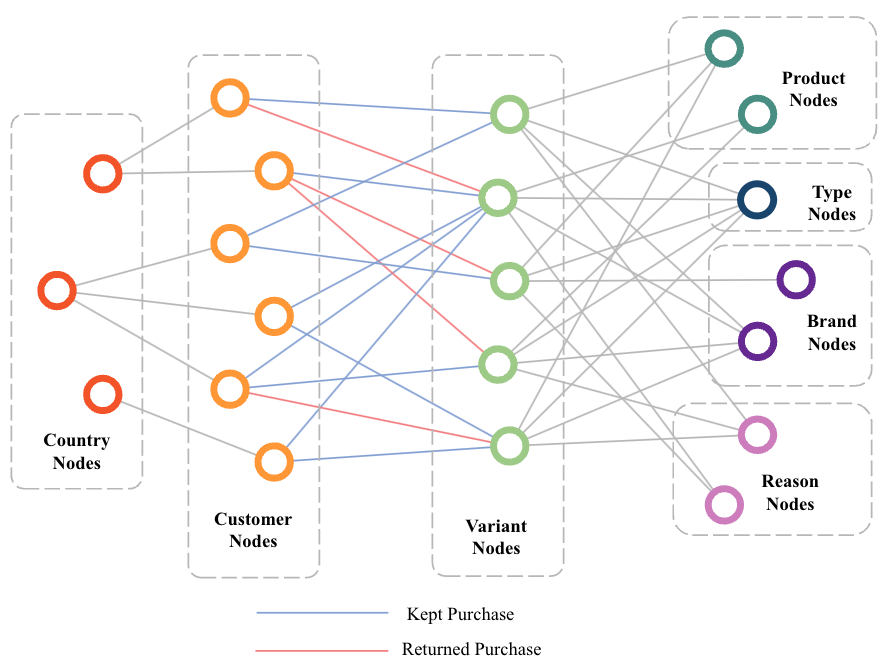}
    \caption{Representation of the richer graph structure contained within the ASOS returns data and how it can be recast into a form better suited to graph representation learning. Virtual nodes are shown for countries, products, product types, brands and return reasons with extra connections added to each customer and product variant node.}
    \label{fig:data_graph_structure}
\end{figure*}
Of particular interest to neural message passing techniques is the inherent graph structure that this dataset holds. In order to apply graph neural networks to data, one must first arrange the data into nodes that contain specific features and edges that link these node instances. This extra potential structure that can be constructed from the ASOS GraphReturns dataset further enhances the modality of the data from the raw structure and node features/correlations discussed above. In Fig.~\ref{fig:data_graph_structure}, we show the data in an undirected heterogeneous graph structure with 5 different edge types linking customers to their shipping countries and product variants to each other and their corresponding brands, product types and top return reasons by defining intermediate virtual nodes in all cases. These virtual nodes can be constructed in multiple ways, however in this paper the virtual nodes contain an averaged set of features for each instance i.e. a product type node will contain the average set of feature values for all products linked to this node.

\section{Methodology}\label{sec:method}
In this section, we present the methodology for a number of example baseline methods applied to the task of predicting customer returns in Sect.~\ref{sec:baseline-results}. The methods considered here aim to provide an early benchmark for future studies involving this dataset. For the graph representation learning based approach, the data is arranged into a highly connected structure with virtual nodes for: customer shipping countries, products, product types, product brands and top return reasons for product variants as described in Fig.~\ref{fig:data_graph_structure}.

We investigate the use of a Logistic Regression, a 2-layer MLP, a Random Forest~\cite{random_forest}, and an XGBoost~\cite{XGboost} classifier trained directly on the raw data (i.e. not arranged into a graph) described in Sect.~\ref{sec:data}. For these models, the customer and product specific features are joined by each labelled purchase link in the data. Further to this, we also investigate a benchmark for a GNN based model trained in conjunction with the same baseline 2-layer MLP as a classifier head. In this case the output of the GNN is the learnt embeddings and the MLP provides a final classification layer for the downstream tasks.

To construct an embedding for an edge $\textbf{e}_{ab}$ between two nodes $a$ and $b$, in general one can perform an operation involving both representations for each node,
\begin{equation}
\textbf{e}_{ab} = \mathcal{O}\left(\textbf{h}_{a}^{(K)}, \textbf{h}_{b}^{(K)}\right).
\end{equation}
where in the case described above, $\mathcal{O}$ is described as a 2-layer MLP classifier which performs the final classification from the output of the GNN.

The output of the MLP classifier head is then the predicted probability for the two class labels (return or no return) which are fed into the cross entropy (CE) loss~\cite{CELoss}:
\begin{equation}
    \mathcal{L}_{\text{CE}} = -\frac{1}{N}\sum_{i=1}^{N}y_{i}\log(p_{i}) + (1-y_{i})\log(1-p_{i})
    \label{eq:CELoss}
\end{equation}
where $N$ is the total number of predictions, $y_{i}$ is the true class label (i.e. 0 or 1 for binary classification) of instance $i$ and $p_{i}$ is the predicted probability for the observation of instance $i$. Here we note that the CE loss takes into account the probability of each classification, whereas the F1-score only considers the final classification label. Therefore it is an important metric to consider when one is interested in robust predictions, as is needed for an effective fashion industry solution for reducing the number of returns.

In order to train the GNN discussed in the following section, an extra step is included into this methodology whereby the purchase events are only trained on if the product variant involved has an average return rate of higher than 80\% or lower than 20\%, in order to provide more robust positive and negative examples of return instances to the GNN. The reason for this is to investigate and avoid issues involving oversmoothing in the representations learnt by the GNN, however all results are quoted on the entire test set with no filtering. The result of this is a dataset with 200,000 purchase events and an average vertex degree for the real nodes of 5 for product variant nodes and 2 for customer nodes.

\section{Experiment Results}\label{sec:baseline-results}
\begin{table}[h]
\centering
\begin{tabular}{@{}ccccc@{}}
\toprule
\multirow{2}{*}{Model} & \multicolumn{4}{c}{Test Scores}        \\ \cmidrule(l){2-5} 
                        & Precision      & Recall         & F1-score       & CE Loss $\mathcal{L}_{\mathrm{CE}}$ \\ \midrule
 Logistic Regression    &  0.723              &  0.726              & 0.725        &  0.602     \\
 Random Forest          &  0.788              &  0.712              &  0.748      &   0.630     \\
 MLP                &   0.870             &   0.656             &  0.748       &   0.582    \\
 XGBoost               &  0.805              &   0.745             &  0.774    &    0.561      \\
 \hline
  \textbf{GNN}                &  \textbf{0.816}              &  \textbf{0.758}            &  \textbf{0.792}    &      \textbf{0.489}    \\\bottomrule
\end{tabular}
\caption{Results for models considered in this section evaluated on the full test data.}
\label{tab:baseline-results}
\end{table}
Table~\ref{tab:baseline-results} displays the precision, recall and F1-scores each model evaluated on the full test dataset (1.5M purchase events). The final hyperparameter values are chosen based on a validation set, randomly and uniformly constructed from 10\% of the training data and are listed as: Logistic Regression ($C = 5.0$, $\mathrm{tol.} = 10^{-4}$), MLP (\# of layers $= 2$, hidden dim. $=128$), Random Forest ($\text{\# of estimators} = 100$, $\text{max. depth} = 6$, $\text{min. samples split} = 2$, $\text{min. samples leaf} = 1$, $\text{max. leaf nodes} = 10$), XGBoost~\cite{XGboost} (booster $=$ gbtree, max. depth $= 4$, $\eta = 0.1$, $\gamma = 1$, min. child weight $= 1$, $\lambda = 2$, objective $=$ Binary Logistic, early stopping rounds $= 5$), GNN (1 GraphSAGE~\cite{GraphSAGE} layer with dim. $= 16$, all aggregations $=$ max. pool, dropout $= 0.2$, normalise $=$ True)\footnote{Any parameters not listed here are left at their default values provided by the packages \texttt{sklearn}~\cite{scikit-learn} (Logistic Regression \& Random Forest), \texttt{xgboost}~\cite{XGboost} (XGBoost), \texttt{PyTorch}~\cite{pytorch} (MLP). and \texttt{PyG}~\cite{pytorch-geo} (GNN).}. For the MLP (16,641 trainable parameters) and GNN (49,665 trainable parameters) models, an Adam optimizer is used with a learning rate of 0.01.

The results in Table~\ref{tab:baseline-results} show a superior performance for a GNN based approach trained on high and low returning examples (described in Section~\ref{sec:method}) across all metrics considered, indicating that a graph-based approach yields a better performing and more robust classification model. For reference, when comparing the same GNN to one trained on all available data, an F1-score of 0.783 was found, suggesting the GNN's performance may suffer from oversmoothing when being trained on less discrete positive and negative examples. Furthermore, as mentioned in Sect.~\ref{sec:method}, the classifier head attached to the GNN is the same MLP model also present in Table~\ref{tab:baseline-results}, therefore supporting the expectation that the graph embeddings from the GNN are able to encode useful information from the data. Table~\ref{tab:baseline-results} also suggests that the GNN's predictions are more robust, based on a lower final CE loss (Equation~\eqref{eq:CELoss}) combined with a higher F1-score evaluated on the test set.

\begin{table*}
\centering
\begin{tabular}{@{}ccccccccc@{}}
\toprule
Model               &  \multicolumn{2}{c}{Country A} & \multicolumn{2}{c}{Country B} & \multicolumn{2}{c}{Country C} & \multicolumn{2}{c}{Country D} \\
\hline
\multicolumn{9}{c}{Trained on all markets}\\
\hline
& F1-score & $\mathcal{L}_{\mathrm{CE}}$ & F1-score & $\mathcal{L}_{\mathrm{CE}}$ & F1-score & $\mathcal{L}_{\mathrm{CE}}$ & F1-score & $\mathcal{L}_{\mathrm{CE}}$ \\
\hline
Logistic Regression & 0.635   &  0.611  & 0.776  & 0.606  & 0.658   & 0.611  & 0.593  & 0.608 \\
Random Forest       & 0.655    & 0.633  & 0.785   & 0.633 & 0.672   & 0.635  & 0.606  & 0.633 \\
MLP & 0.680    & 0.527  & 0.792  & 0.527  & 0.691   & 0.528  & 0.626 & 0.518 \\
XGBoost       & 0.731    &  0.556 & 0.806   & 0.567 & 0.717   & 0.567  & 0.664  & 0.561 \\
\midrule
GNN          &    \textbf{0.757}   &  \textbf{0.436} &  \textbf{0.821}  & \textbf{0.487} & \textbf{0.744}  &  \textbf{0.485}  &  \textbf{0.732} & \textbf{0.494} \\
\hline
\end{tabular}

\vspace{0.2cm}

\begin{tabular}{@{}ccccccccc@{}}
\toprule
Model               &  \multicolumn{2}{c}{Country E} & \multicolumn{2}{c}{Country F} & \multicolumn{2}{c}{Country G}    & \multicolumn{2}{c}{Country H} \\
\hline
\multicolumn{9}{c}{Trained on all markets}\\
\hline
& F1-score & $\mathcal{L}_{\mathrm{CE}}$ & F1-score & $\mathcal{L}_{\mathrm{CE}}$ & F1-score & $\mathcal{L}_{\mathrm{CE}}$ & F1-score & $\mathcal{L}_{\mathrm{CE}}$ \\
\hline
Logistic Regression  & 0.812  & 0.591 & 0.729   &  0.618  & 0.673   & 0.605  & 0.671  & 0.610 \\
Random Forest       & 0.817  & 0.624 & 0.745   &  0.638   & 0.717   & 0.630  & 0.683   & 0.636 \\
MLP & 0.819  & 0.514 & 0.754   &  0.542  & 0.727   & 0.520  & 0.696  & 0.528 \\
XGBoost       & 0.827  & 0.561 & 0.772   &  0.573   & 0.751   &  0.561 & 0.728  &  0.563\\
\midrule
GNN          &   \textbf{0.842}  & \textbf{0.487} &   \textbf{0.801}   & \textbf{0.500}  &  \textbf{0.774}  &  \textbf{0.489}  &   \textbf{0.744} & \textbf{0.505} \\
\hline
\end{tabular}
\caption{Summary of F1-scores and CE losses ($\mathcal{L}_{\mathrm{CE}}$) evaluated on the test data for each individual country market. In these results we use a GNN model with 1 SAGEGraph layer (dim. = 16) trained with all extra nodes considered from Sect.~\ref{sec:method}.}
\label{tab:country-results-with-nodes}
\end{table*}
Table~\ref{tab:country-results-with-nodes} displays the F1-scores evaluated on the test set for individual country markets. In all country instances, the GNN based approach obtains a superior F1-score to all other models considered. When comparing the results in these tables with the correlations discussed in Fig.~\ref{fig:data_corr_example} one can observe that those countries with higher correlations to a particular return label (1 or 0) are among the top performing F1-scores in Table~\ref{tab:country-results-with-nodes}.

Single market results are of particular interest to the wider e-commerce fashion industry in order to understand how to deliver the best service to customers and products across different individual markets. The ability to obtain results such as these are an important and unique feature in the novel ASOS GraphReturns dataset as it facilitates a level of understanding into how an ML model is performing across different areas and identify it's weaknesses. Note that a similar analysis can be done for different brands or product types.

\section{Conclusion}\label{sec:conclusion}
In this work we have presented a novel dataset to inspire new directions in fashion retail research. This dataset is particularly suited to graph representation learning techniques and exhibits a naturally rich geometrical structure. 

The baseline models which have been presented here to provide an early benchmark trained on the presented data support the claim that a GNN based approach achieves a higher yield over the metrics considered. The best performing model is a GNN model described in Sect.~\ref{sec:method} and \ref{sec:baseline-results} which obtained a final F1-score of 0.792 and a test CE loss score of 0.489 when evaluated on the test set. These results are an improvement from the next best performing model (2\% higher F1-score and 6\% lower CE loss) indicating the potential for graph based methods on this naturally graph structured data. Of particular interest for e-commerce companies is the level of confidence when making a prediction which will affect the likelihood of a customer being notified by the prediction. Therefore the final test CE loss value for the GNN being lower than other models implies that overall the GNN is likely more confident about its classifications than the other non-graph based approaches. In order to reinforce this point, a future analysis of these predictions could include the investigation of calibrated probabilities as in~\cite{Guo:calibration}.

As discussed, our primary goal is to provide a novel dataset to facilitate future research studies in fashion retail. This data is presented with labeled purchase links between customers and product variants which can be used in a supervised learning setting (as in Sect.~\ref{sec:baseline-results}). However due to the graph structure of this data, it is possible to also use this data in the unsupervised setting with a wider range of transformer based models. Finally we wish to highlight the potential application of this dataset to advancements in recommendation systems. With the definite labels provided in this dataset which label a return, a future research direction would be investigating the universality of the GNN embeddings and how these translate into new recommendation systems for sustainable fashion.

\bibliographystyle{ACM-Reference-Format}
\bibliography{references}

\end{document}